\pdfoutput=1

\documentclass[11pt]{article}

\usepackage[final]{acl}

\usepackage{times}
\usepackage{latexsym}
\usepackage{amsmath}

\usepackage[T1]{fontenc}

\usepackage[utf8]{inputenc}

\usepackage{microtype}

\usepackage{inconsolata}

\usepackage{graphicx}

\newcommand{\dist}{\mathop{\mathrm{dist}}}

\title{Persistent Homology of Topic Networks for the Prediction of Reader Curiosity}

\author{
 \textbf{Manuel D. S. Hopp\textsuperscript{1}},
 \textbf{Vincent Labatut\textsuperscript{2}},
 \textbf{Arthur Amalvy\textsuperscript{2}},
 \textbf{Richard Dufour\textsuperscript{3}},\\
 \textbf{Hannah Stone\textsuperscript{4}},
 \textbf{Hayley Jach\textsuperscript{5,1}},
 \textbf{Kou Murayama\textsuperscript{1}}
\\
\\
 \textsuperscript{1}Hector Research Institute of Education Sciences and Psychology, University of Tübingen, Germany\\
 \href{mailto:manuel.hopp@uni-tuebingen.de, k.murayama@uni-tuebingen.de}{\texttt{\{firstname.lastname\}@uni-tuebingen.de}},\\
 \textsuperscript{2}Laboratoire Informatique d'Avignon -- LIA UPR 4128, Avignon Université, France\\
 \href{mailto:vincent.labatut@univ-avignon.fr, arthur.amalvy@univ-avignon.fr}{\texttt{\{firstname.lastname\}@univ-avignon.fr}},\\
 \textsuperscript{3}Laboratoire des Sciences du Numérique de Nantes -- LS2N UMR 6004, Nantes Université, France\\
 \href{mailto:richard.dufour@univ-nantes.fr}{\texttt{richard.dufour@univ-nantes.fr}},\\
 \textsuperscript{4}Emerald Publishing, Reading, United Kingdom\\
 \href{mailto:hannah.j.stone@reading.ac.uk}{\texttt{hannah.j.stone@reading.ac.uk}},\\
 \textsuperscript{5}Melbourne School of Psychological Sciences, The University of Melbourne, Australia\\
 \href{mailto:hayley.jach@unimelb.edu.au}{\texttt{hayley.jach@unimelb.edu.au}}
\\
}

\begin{document}
\maketitle
\begin{abstract}
Reader curiosity, the drive to seek information, is crucial for textual engagement, yet remains relatively underexplored in NLP. Building on Loewenstein's Information Gap Theory, we introduce a framework that models reader curiosity by quantifying semantic information gaps within a text's semantic structure. Our approach leverages BERTopic-inspired topic modeling and persistent homology to analyze the evolving topology (connected components, cycles, voids) of a dynamic semantic network derived from text segments, treating these features as proxies for information gaps. To empirically evaluate this pipeline, we collect reader curiosity ratings from participants ($n = 49$) as they read S. Collins's ``The Hunger Games'' novel. We then use the topological features from our pipeline as independent variables to predict these ratings, and experimentally show that they significantly improve curiosity prediction compared to a baseline model (73\% vs. 30\% explained deviance), validating our approach. This pipeline offers a new computational method for analyzing text structure and its relation to reader engagement. 
\end{abstract}
\textcolor{red}{\textit{This version updates the ACL 2025 article with extended appendices, corrected topological definitions, and minor errata.}}

\section{Introduction}
\label{sec:Intro}
Reader curiosity refers to the cognitive and affective drive that motivates individuals to seek additional information while reading. In the context of textual engagement, this can manifest as a reader's urge to continue reading, explore related topics, or seek clarifications~\cite{text_Schieferle1999}. While curiosity is often studied in educational psychology, its computational modeling in natural language processing (NLP) remains relatively underexplored. 
Existing approaches to modeling reader engagement often rely on linguistic features, e.g., sentiment analysis, readability scores~\cite{NLPsenti_Sotirakou2021}, or word-level analyses~\cite{NLPlingu_Berger2023, NLPemotion_Maslej2021, NLPemotion_Dvir2023}. These methods, while valuable, primarily focus on surface-level characteristics and often fail to capture the broader semantic structure, narrative flow, and, crucially, the information gaps that stimulate curiosity. Related work leveraging knowledge graphs, while providing a richer semantic representation, does not explicitly model the reader's evolving understanding and points of information need~\cite{Knowlgreview_Abu2024}. 

Building on Loewenstein's \textit{Information Gap Theory}~\cite{Loewenstein1994}---where curiosity stems from recognizing a difference between known and desired information---we introduce a framework that models reader curiosity by quantifying semantic information gaps within a text's semantic structure. Unlike prior work focused on micro-level textual features, our approach adopts a macro-level, cognitive perspective, operationalizing the concept of \textit{plot holes}, or \textit{information gaps}, to predict reader engagement. We hypothesize that these gaps, representing areas of missing connections or coherence in the textual flow, act as intrinsic motivators. Furthermore, we operationalize surprise, a key driver of curiosity, by measuring the dynamic shifts and transformations in these information gaps throughout the text. 

To realize this framework, we introduce a pipeline leveraging Topological Data Analysis (TDA). This pipeline integrates recent topic modeling (building upon BERTopic~\cite{BERTopic_Grootendorst2022}) to extract key topics, constructs a dynamic topic network representing the flow of these topics~\cite{Timeskelton_Zhu2013}, and applies TDA, specifically Persistent Homology~\cite{TDA_Munch2017}, to identify and quantify topological cavities within this network~\cite{wiki_Patankar2023, wiki_Zhou2024}. These cavities---disconnected components, cycles, and voids---represent information gaps. We further employ Wasserstein and Bottleneck distances to measure the evolution of these gaps, capturing the element of surprise. This study explores the potential of this novel approach of characterizing topological features of a text.

To demonstrate feasibility, we conducted a pilot study as a proof-of-concept using S. Collins's ``The Hunger Games'' \citep{HungerGames_Collins2021} novel. Participants na\"{\i}ve to both the book and its movie adaptation provided chapter-wise curiosity ratings, enabling an initial analysis of curiosity dynamics in response to information gaps.


Our main contributions are fourfold:
\begin{enumerate}
    \item \textbf{Pipeline:} We designed a pipeline for the modeling of textual information gaps, integrating topic modeling and TDA.
    \item \textbf{Engagement data:} We conducted a survey in order to obtain reader engagement data for ``The Hunger Games'' novel. 
    \item \textbf{Experimental validation:} We leveraged the survey data to evaluate our approach empirically. 
    \item \textbf{Interdisciplinary approach:} Our method connects topic modeling and TDA with theories from motivational psychology, facilitating further interdisciplinary research.
\end{enumerate}
We share our source code and data at \url{https://github.com/mds-hopp/pers_homol_data}.

The rest of this article is organized as follows. In Section~\ref{sec:Related}, we review the literature directly related to our work. In Section~\ref{sec:Methods}, we present our narrative modeling pipeline. Section~\ref{sec:Setup} describes our experimental setup, while our results are presented and discussed in Section~\ref{sec:Results}. Finally, we review the salient points of our work and its perspectives in Section~\ref{sec:Conclu}.

\section{Related Work}
\label{sec:Related}
This work builds upon and addresses limitations in computational reader engagement models, graph-based and topological methods in NLP, and network-based approaches to curiosity and exploration.

\paragraph{Computational Models of Reader Engagement} Predicting user engagement is an important NLP task, often approached through text feature analysis. There are many features that are related to reader engagement. Early methods relied on surface characteristics like sentiment and readability~\cite{NLPsenti_Sotirakou2021}. Other approaches often relied on classical bag-of-words representations for word-level analysis~\cite{NLPemotion_Maslej2021, NLPemotion_Dvir2023}. However, these methods, while useful, largely neglect semantic structure and cognitive processes. Other work incorporates cognitive aspects, highlighting uncertainty~\cite{NLPlingu_Berger2023} and semantic cohesion~\cite{SemanticCohesion_Ward2008}. Yet, these typically operate at the word or sentence level, failing to model the reader's \textit{evolving} information state---critical to theories like Loewenstein's Information Gap Theory~\cite{Loewenstein1994}. Our pipeline directly addresses this, modeling the \textit{dynamic} evolution of semantic information gaps. The use of text embeddings from multilingual large language models also opens the door to potential cross-cultural work on the topic, e.g.~\cite{wiki_Zhou2024}.

\paragraph{Graph \& Topological Methods in NLP} The drive for higher-level text understanding has increased the use of graph representations in NLP. Knowledge graphs enhance tasks like question answering~\cite{Knowlgreview_Abu2024}, and graph-based retrieval augmented generation (RAG) methods, such as GraphRAG~\cite{graphrag_Han2025} and LightRAG~\cite{lightrag_guo2024}, leverage relational structure. Topic modeling also benefits from graph approaches. Traditional methods like Latent Dirichlet Allocation (LDA) are complemented by approaches using bipartite networks and community detection~\cite{NetworkApproachTopic_Gerlach2018}, and embedding-based methods like BERTopic~\cite{BERTopic_Grootendorst2022} or a combination of both embeddings and networks~\cite{ConstructionKnowledge_Cao2019}, offering richer semantic representations. However, these generally do not analyze the \textit{topological} structure of the resulting networks. 

Topological Data Analysis (TDA), particularly Persistent Homology, provides tools to analyze data shape, including networks. A recent review shows TDA's growing interest in NLP~\cite{TDANLPrev_Uchendu2024}. \citet{math_network_Christianson2020} used TDA to identify knowledge gaps in math textbooks, and \citet{logical_holes_Tymochko2021} to capture logical holes in abstracts. Critically, existing NLP applications of TDA primarily focus on \textit{static} text representations. Our work significantly extends this, applying persistent homology to a \textit{dynamic} topic network (inspired by the work of \citet{Timeskelton_Zhu2013}), tracking the evolution of topological features (specifically, cavities) over time. This dynamic aspect is crucial for modeling changing reader information gaps.

\paragraph{Engagement and Graph \& Topological Methods} Beyond NLP, network science has modeled text structure to understand its implications to learning and cognition, including curiosity-driven exploration~\cite{wiki_Zhou2024, wiki_Patankar2023}. \citet{wiki_Patankar2023} tracked structural changes in time-varying graphs using persistent homology, conceptually related to our approach. However, their focus was on general graph dynamics, not reader engagement. We bridge this gap, connecting network models of exploration with cognitive theories of curiosity, specifically the Information Gap Theory, applying them to model reader engagement in NLP. Operationalizing information gaps as topological cavities in a dynamically evolving semantic network provides a novel, quantifiable measure of reader curiosity, directly linking computational methods and psychological theories. Moreover, our methodology targets continuous sequential texts (like narratives) to analyze intrinsic semantic evolution and its link to reader curiosity. To analyze how semantic evolution in continuous narratives relates to reader curiosity, we required paired text-curiosity annotations. As our literature review confirmed no dataset appropriate for tracking narrative semantic shifts exists, we collected our own for this study.

\section{Methods for Narrative Modeling}
\label{sec:Methods}
\begin{figure}[htb!]
    \centering
    \includegraphics[width=.95\linewidth]{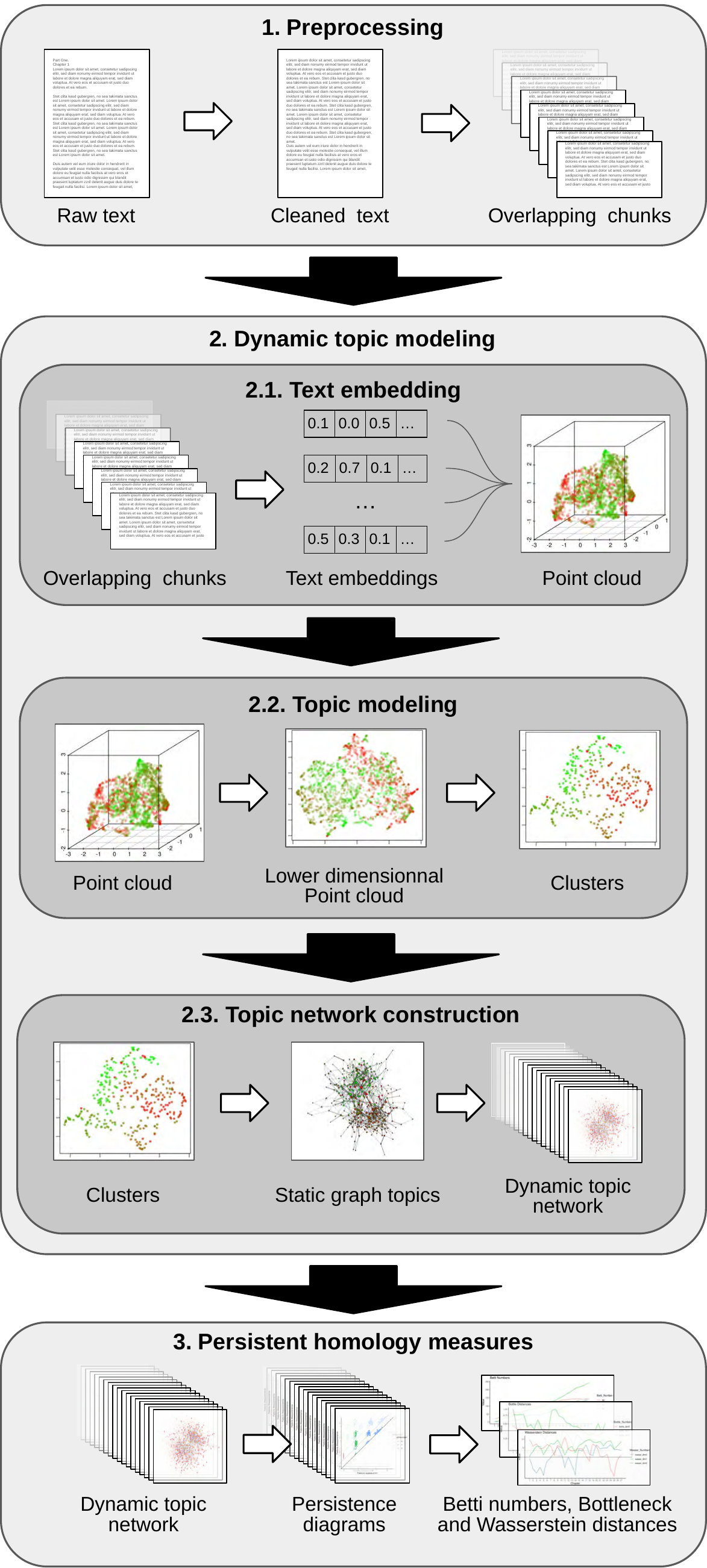}
    \caption{Pipeline from preprocessing (top) to persistent homology measures (bottom).}
    \label{fig:pipeline}
\end{figure}

We employ a combination of NLP, network analysis, and TDA to model narrative structure and its evolution, building on and extending previous work. Our analysis proceeds in three key stages: 
\begin{enumerate}
    \item \textbf{Preprocessing}. We obtain textual data and segment it using a sliding window approach.
    \item \textbf{Dynamic Topic Modeling}. A dynamic topic network is built to represent the evolving thematic structure of the text. Vertices represent topics, and edges connect topics appearing in consecutive text segments, with edge weights reflecting semantic similarity.
    \item \textbf{Topological Feature Extraction via Persistent Homology}. Persistent homology quantifies the network's evolving topological features representing information gaps of the topic network.
\end{enumerate}
An overview of our suggested pipeline (from text preprocessing to the topological feature extraction) is depicted in Figure~\ref{fig:pipeline}. Details regarding the data and models we employed are located in Section~\ref{sec:Setup}.

\subsection{Dataset and Preprocessing}
Our approach works with either textual or multimodal data (e.g., video), adaptable to the chosen embedding model. After data collection, the next step involves cleaning and segmenting the textual or multimodal data into smaller units. A key aspect is the collection of user engagement ratings periodically throughout content consumption. The human data allows us to validate the persistent homology measures as reader engagement.

\subsection{Dynamic Topic Network}
To capture the evolving thematic structure of the used texts, we construct a dynamic topic network. Each vertex in this network represents a topic and edges connect topics that occur across consecutive chunks, reflecting the narrative flow of the story. Each edge has a weight corresponding to the cosine similarity of the dyadic topics in the text embedding space. 

\subsubsection{Topic Modeling}
We employ a pipeline inspired by the BERTopic approach for topic modeling~\cite{BERTopic_Grootendorst2022}. It consists of the following three stages: 
\begin{enumerate}
    \item \textbf{Text Embedding}: Text segments are embedded into a high-dimensional vector space using the transformer-based embedding model voyage-3-large~\cite{VoyageAI2025}. These embeddings capture the semantic meaning of the chunks, with similar chunks having closer embeddings.
    \item \textbf{Dimensionality Reduction}: As a preprocessing step for clustering, we reduce the high-dimensional embedding vectors using UMAP (Uniform Manifold Approximation and Projection; \citealp{UMAP_Mcinnes2020}) following the BERTopic approach~\cite{BERTopic_Grootendorst2022} for improving cluster quality with high-dimensional data~\cite{UMAPcluster_Asyaky2021, UMAPcluster_Allaoui2020}.
    \item \textbf{Clustering}: We use HDBSCAN (Hierarchical Density-Based Spatial Clustering of Applications with Noise)~\cite{hdbscan_Campello2013}, which identifies non-convex clusters of varying densities and handles noise explicitly. Furthermore, HDBSCAN automatically determines the cluster count, avoiding manual parameter tuning and potential bias. Its strong empirical performance further justifies its use~\cite{hdbscan_scikit2025, UMAPcluster_Asyaky2021, hdbscancomparing_Mcinnes2016}. Each topic is represented by the weighted centroid of its constituent text chunk embeddings, using the cluster probability as the weight. 
\end{enumerate}

\subsubsection{Dynamic Topic Network Construction}
\label{dyntopnet}
The dynamic topic network is built incrementally, based on the measurement points of the user engagement sampling. 

\begin{itemize}
    \item \textbf{Vertices:} Each vertex in the network represents one topic.
    \item \textbf{Edges:} Undirected edges are created between topic vertices appearing in consecutive text chunks. This captures the sequential flow of topics throughout the narrative, as suggested by~\citet{Timeskelton_Zhu2013} and applied in studies such as~\citet{wiki_Patankar2023}. 
    \item \textbf{Edge Weights:} The weight of each edge is determined by the cosine similarity between the embedding vectors of the connected topics in the original embedding space. Higher cosine similarity results in a stronger connection. 
    \item \textbf{Network Series:} We segment the narrative based on user engagement rating points. Each such segment is represented by a distinct static topic graph built upon the topics and relationships occurring in the narrative up to this point. This sequence of static graphs forms a cumulative dynamic network: The first graph represents only the topics and relationships up to the first user engagement rating point, the second graph contains these topics and relations plus those occurring in the second segment, and so on. The last graph in the sequence represents all topics and relationships.
\end{itemize}

\subsection{Persistent Homology Measures}
\begin{figure}[htb!]
    \centering
    \includegraphics[width=.8\linewidth]{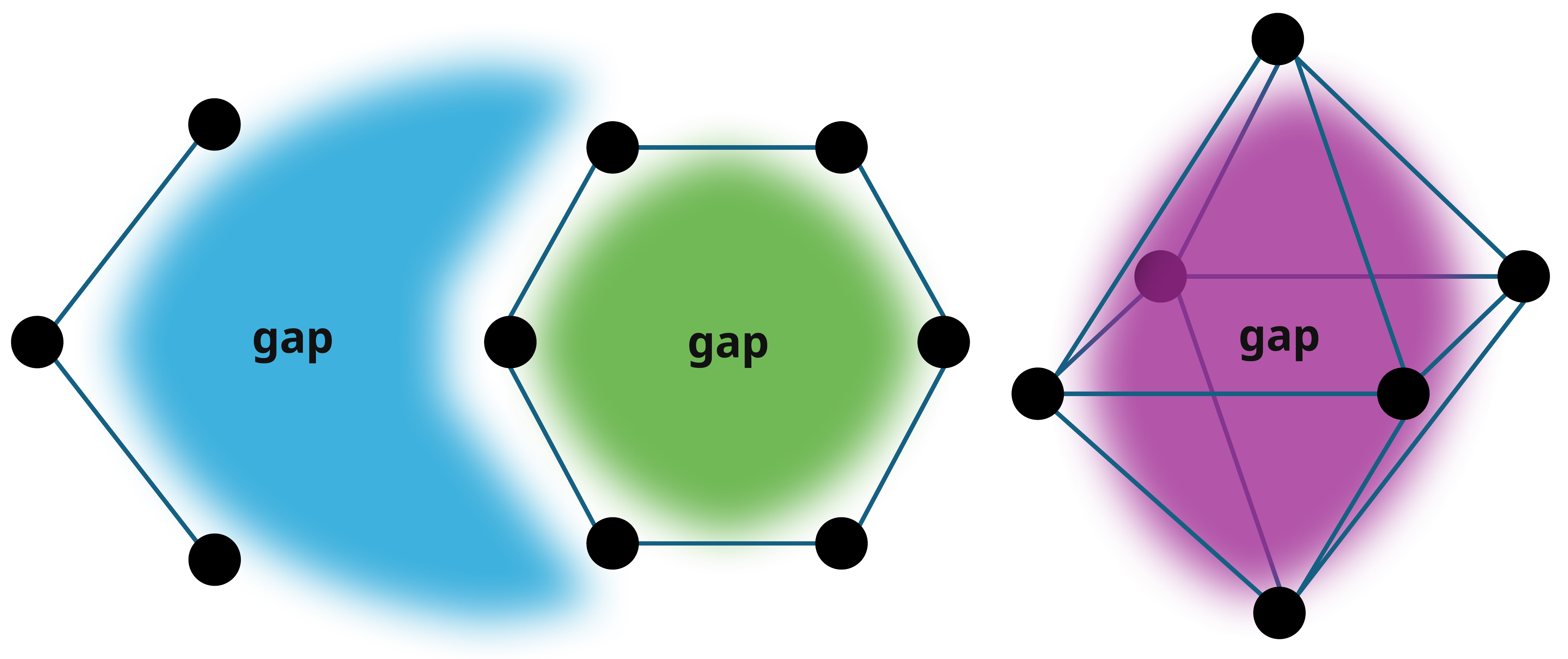}
    \caption{Identified topological features of the topic network: Components, cycles/ loops, and voids. Adapted from~\citet{wiki_Patankar2023}.}
    \label{fig:gaps}
\end{figure}

To evaluate the gaps in information flow within our dynamic topic networks, we employ persistent homology, a method derived from topological data analysis. In essence, persistent homology allows us to detect and monitor the evolution of specific topological features within a network (in our case, the topic network using cosine similarity-derived distances)---namely, connected components, cycles (loops), and voids (enclosed empty spaces)~\cite{TDA_Moroni2021, TDA_Munch2017}---that we compare in Figure~\ref{fig:gaps}. For a brief, non-technical introduction to Topological Data Analysis and Persistent Homology, please refer to Appendix~\ref{sec:IntrTDA}.

\begin{figure}[htb!]
    \centering
    \includegraphics[width=.9\linewidth]{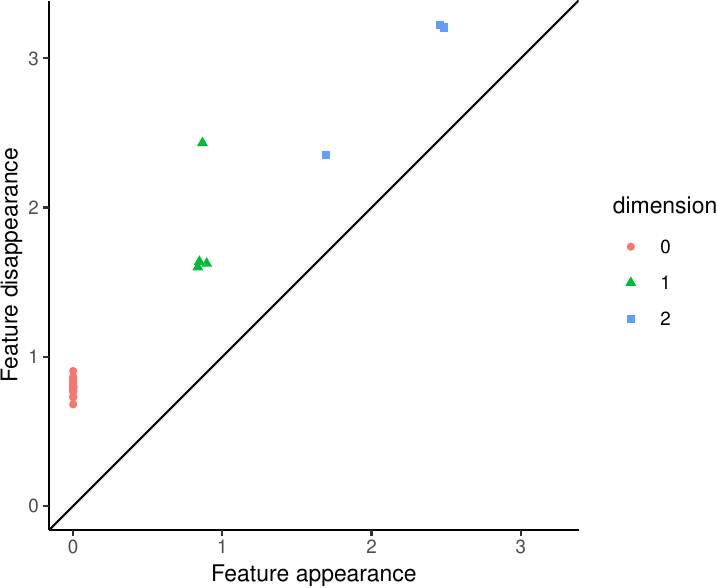}
    \caption{Persistence diagram of the first chapter of our dataset.}
    \label{fig:persist}
\end{figure}

The core of persistent homology lies in examining how a network's structure evolves as a function of a filtration parameter, denoted by $\epsilon$. Intuitively, $\epsilon$ represents a proximity threshold. As $\epsilon$ increases, connections (edges, and in more general cases, higher-dimensional counterparts called simplices) are progressively added between vertices that are closer than the given $\epsilon$. This process generates a nested sequence of simplicial complexes, each representing the network's structure at a specific proximity level. From this sequence, persistent homology tracks the ``birth'' (emergence) and ``death'' (merging or disappearance) of topological features across dimensions as $\epsilon$ increases: 0-dimensional connected components, 1-dimensional cycles (loops), and 2-dimensional voids (enclosed empty spaces). The results are summarized in a persistence diagram, which plots each finite persistence interval as a point $(b, d)$, where $b$ signifies the $\epsilon$ value at which the feature is born, and $d$ represents the $\epsilon$ value at which it dies. Rather than counting instantaneous Betti numbers at a single threshold $\epsilon$, we record the total count of finite persistence intervals born across the filtration for each dimension, denoted $\beta_0$, $\beta_1$, and $\beta_2$. Figure~\ref{fig:persist} shows the persistence diagram for the first chapter of our dataset, revealing one highly persistent feature in dimension 1 (loop).

\section{Experimental Setup}
\label{sec:Setup}

\subsection{Dataset and Preprocessing}
\label{sec:SetupPrepr}
\subsubsection{Dataset} The primary dataset for this pilot study consists of the young adult novel ``The Hunger Games'' by \citet{HungerGames_Collins2021} and corresponding reader engagement data ($n = 76$ participants; mean age = 41.4, 56\% female, 100\% UK residence) collected through the online platform Prolific\footnote{\url{https://www.prolific.com}}. Participants were compensated in line with Prolific's recommended minimum 7.50 British Pounds per hour.  Prior to reading, participants indicated their familiarity with the book and its movie adaptation. During reading, participants provided continuous self-reported ratings (0--100) on multiple engagement dimensions after each chapter. For this analysis, we focus specifically on curiosity ratings (``I was curious about this chapter'') from readers unfamiliar with either the movie or book ($n = 49$). As we wished to assess the general pattern of curiosity responses (i.e., not individual reader differences, which we could not examine in this work), we computed the mean curiosity rating across participants for each chapter. Inter-Rater Reliability for mean chapter curiosity, calculated via mixed-effects models~\cite{Shrout1979} (cf. Appendix~\ref{sec:Icc}), is .71 (moderate), showing reasonable consistency across chapters. 

\subsubsection{Preprocessing}
We apply a two-step text preprocessing. First, we remove part titles (e.g., ``Part One''), chapter titles (e.g., ``The Tributes''), chapter numbers (e.g., ``Chapter 1''), and empty lines. Second, we employ a sliding window approach to create text segments, using windows of 5 sentences with a 2-sentence overlap. This results in 2,656 text segments. The median chunk length is 60 words (MAD = 20.76, range = 13--155 words). 

\subsection{Topic Modeling}
\paragraph{Text Embedding:} We embed text segments into a 1,024-dimensional vector space using the Voyage-AI voyage-3-large transformer-based embedding model~\cite{VoyageAI2025} via the VoyageAI API, accessed through Python. We select transformer-based models, and especially this one, due to its state-of-the-art performance on the Massive Text Embedding Benchmark (MTEB)~\cite{MTEP_Muennighoff2022}, indicating its ability to capture nuanced semantic relationships~\cite{embedding_Morris2023, Embeddings_Yu2024}. 

\paragraph{Dimensionality Reduction:} As a preprocessing step for clustering, we reduce the 1,024-dimensional embedding vectors to 32 dimensions using UMAP~\cite{UMAP_Mcinnes2020}. The UMAP parameters are set to the cosine similarity metric and 15 nearest neighbors to ensure the preservation of both global and local structures in the lower-dimensional representation. 

\paragraph{Clustering:} Clustering is performed using HDBSCAN~\cite{hdbscan_Campello2013} with a minimum cluster size of 3 data points, for fine-grained results. This results in the identification of 302 topics. Embeddings identified as noise ($n = 717$, 27\%) are excluded from further analysis. 

These steps were implemented in R~\cite{R_Rteam2024} using the \texttt{uwot} package~\cite{UWOT_Melcille2024} for the UMAP implementation, and \texttt{dbscan}~\cite{dbscan_Hahsler2019} for hierarchical density-based clustering.

\subsection{Persistent Homology Measures}
\label{sec:SetupPersist}
We first create a series of 27 static topic graphs (as outlined in Section~\ref{dyntopnet}) and then use a Vietoris--Rips filtration to construct a simplicial complex from each graph~\cite{Sheehy2012}. The filtration is based on the edge weights (cosine similarity), with simplices (edges, triangles, etc.) added as the distance threshold increases. 
From the resulting persistence diagrams, we extract the following topological features:
\begin{itemize}
    \item \textbf{Persistence Feature Counts}: $\beta_0$, $\beta_1$, and $\beta_2$ denote the total counts of finite persistence intervals (homological generators) born across the filtration up to the graph diameter for dimension 0 (connected components), dimension 1 (cycles/loops), and dimension 2 (voids/cavities), respectively. They quantify the network's overall capacity to form multi-scale information gaps. (Because \texttt{TDAstats} omits the single infinite-lifetime component, $\beta_0$ equals $|V| - 1$ for each connected cumulative graph.)
    \item \textbf{Bottleneck Distance}: $\dist_B$ measures the maximum difference between the persistence diagrams of consecutive graphs (i.e., between graphs representing chapters $n$ and $n + 1$). A large bottleneck distance indicates a significant singular structural change, such as the creation or filling of a large void.
    \item \textbf{Wasserstein Distance}: $\dist_W$ measures the total ($L_1$ optimal-transport) difference between the persistence diagrams of consecutive graphs. A large Wasserstein distance indicates a substantial aggregate structural shift across all features, such as widespread changes in cycle or void sizes.
\end{itemize}

We detrend the data using residuals from a linear model fitted to the chapter index and winsorize it due to the presence of outliers. Specifically, for each feature, values below the 2.5\textsuperscript{th} percentile are capped at the 2.5\textsuperscript{th} percentile, and values above the 97.5\textsuperscript{th} percentile are capped at the 97.5\textsuperscript{th} percentile. 

Network analysis is handled by the \texttt{igraph} package~\cite{igraph_Csardi2006} with \texttt{qgraph}~\cite{graph_Epskamp2012}, in R. Persistent homology and related distance measures are computed using the R packages \texttt{TDA}~\cite{TDA_Brittany2024} and \texttt{TDAstats}~\cite{TDAstats_Wadhwa2018}. Single-threaded persistent homology calculations take approximately 100 seconds on an AMD Ryzen 7 7840U (64 GB RAM).

\subsection{Generalized Additive Model}
\label{sec:SetupStats}
To investigate the relationship between the topological features extracted from the text and the readers' reported curiosity, we employ a Generalized Additive Model (GAM) using the R package \texttt{mgcv}~\cite{GAM_Wood2011}. We choose GAMs for two key reasons: their ability to capture non-linear relationships and their robust options for addressing overfitting, which is crucial given our limited sample size.

We assess topological features' unique contribution to explaining variance in reader curiosity by comparing a \textit{Null Model} (control variables: novel topics per chapter, chapter index) and a \textit{Full Model} (control variables + topological features: persistence feature counts, Wasserstein distances, Bottleneck distances).

Due to the limited number of observations ($n = 27$ chapters), our primary goal is to explore the explained deviance of the model using topological features, rather than making definitive claims about the precise functional form of the relationships. The dependent variable (DV) is the mean curiosity reported per chapter (based on $n = 49$ observations per chapter). Predictor variables (IVs) include detrended persistence feature counts ($\beta_0, \beta_1, \beta_2$), Wasserstein and Bottleneck distances between chapters, and, as control variables, the number of novel topics per chapter and the chapter index number itself. 

We use cubic regression splines for all smooth functions, setting the basis dimension $k$ to 4 for all smooth terms based on GAM diagnostics, as recommended by~\citet[Section~5.9]{GAM_Wood2017}. To avoid overfitting, we use Restricted Maximum Likelihood (REML) for parameter estimation with an additional penalty term ($\gamma = 1$) during model fitting.
To assess the significance of the explained deviance and $R^2$, we employ a permutation test with 1,000 iterations, where we permute the values of the mean curiosity across chapters and refit the model.

\section{Results}
\label{sec:Results}
\subsection{Distribution of the Detected Topics}
\begin{figure}[htb!]
    \centering
    \includegraphics[width=1\linewidth]{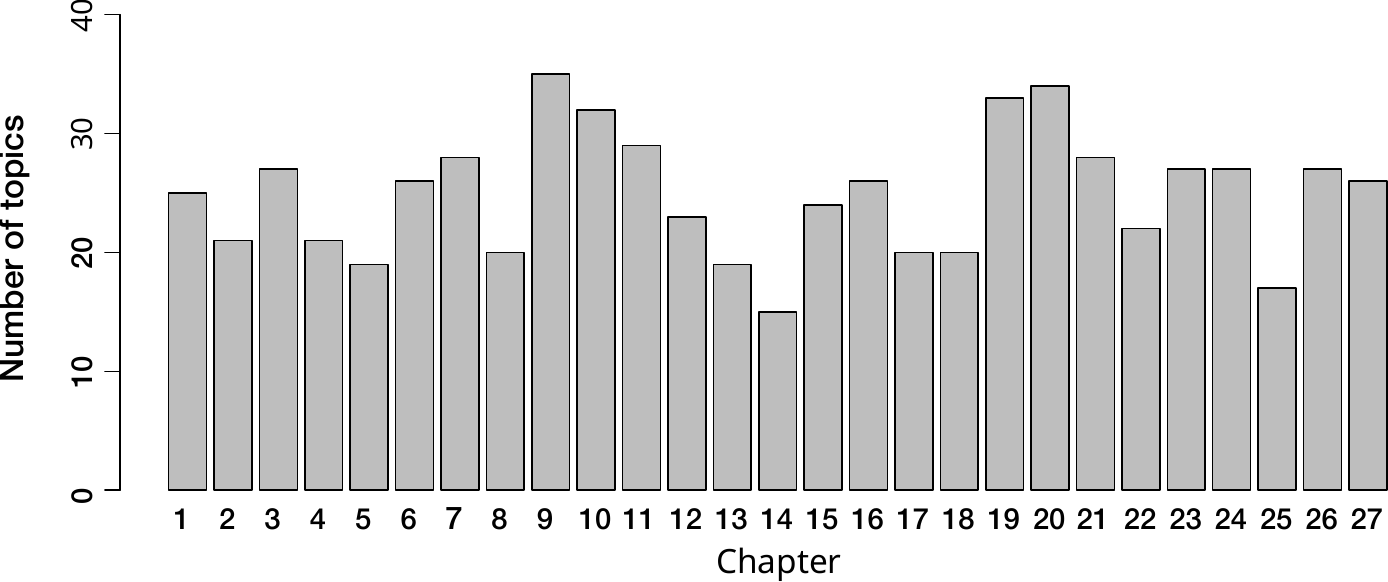}\\
    \includegraphics[width=1\linewidth]{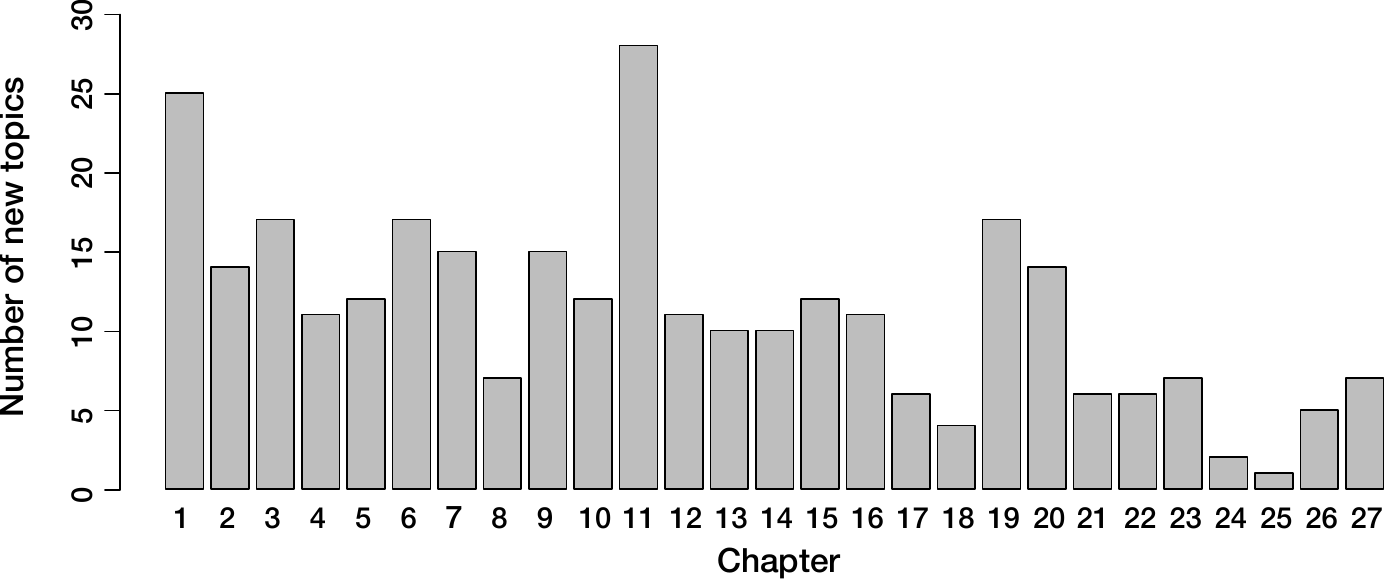}
    \caption{Total number of topics by chapter (top) and number of new topics by chapter (bottom). The $x$-axis represents chapter numbers, and the $y$-axis shows the number of topics and new topics identified within that chapter, respectively.}
    \label{fig:TopicStats}
\end{figure}

HDBSCAN identifies 302 distinct clusters, i.e., topics, over the entire text. A comprehensive summary of these topics, generated using DeepSeek V3~\cite{deepseek_2024}, is provided on the online repository hosting our source code. 
Figure~\ref{fig:TopicStats} shows the distribution of topics across chapters and the number of new topics by chapter, respectively. On average, each chapter contains 25 topics (SD = 5, range: 15--35). Chapter~9 exhibits the highest number of topics, while Chapter~11 contains the largest number of newly introduced topics ($n = 28$). In contrast, Chapter~14 has the fewest total topics, and Chapter~25 introduces the fewest new topics ($n = 1$). 

\begin{figure}[htb!]
    \centering
    \includegraphics[width=.9\linewidth]{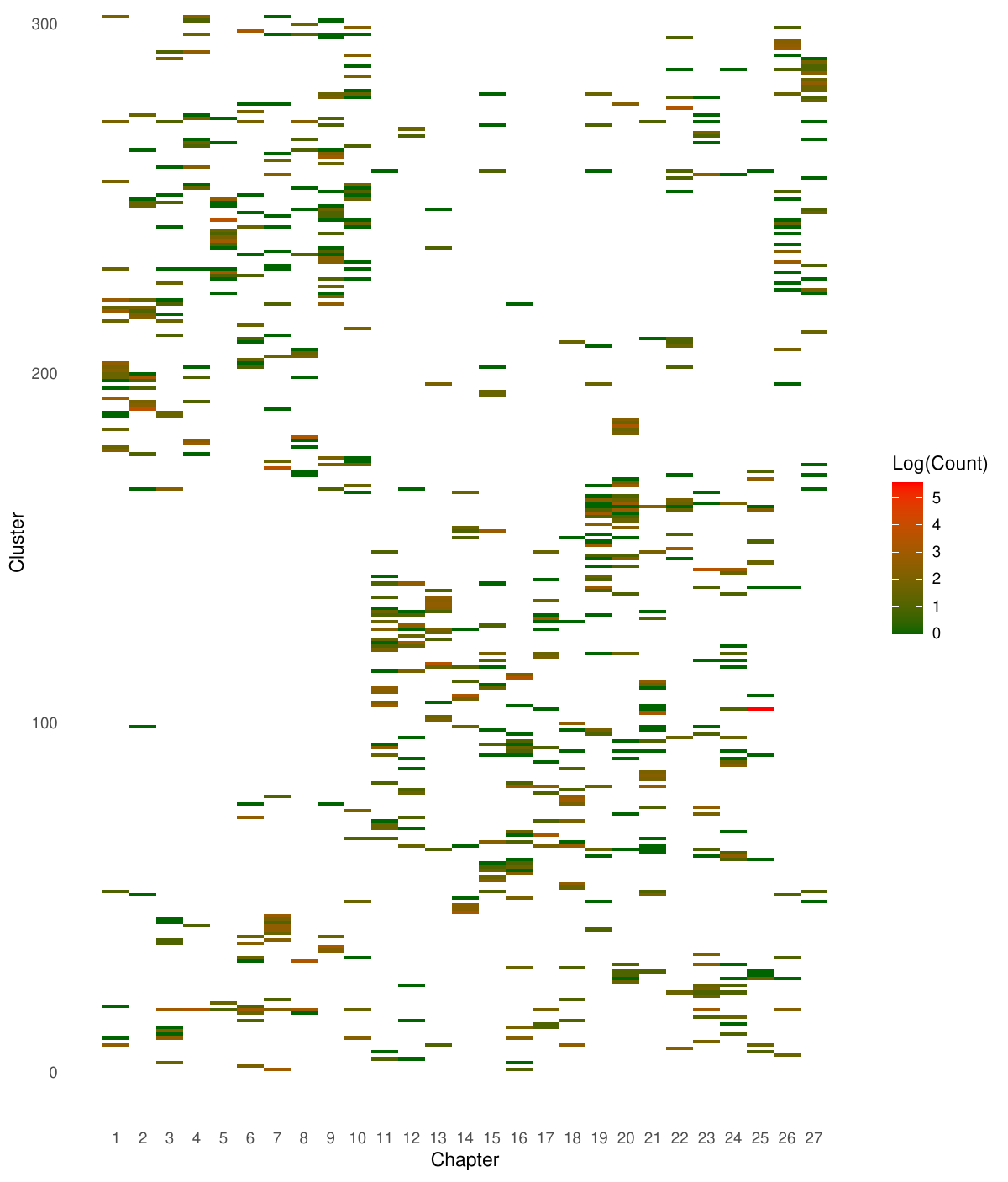}
    \caption{Distribution of topics across chapters. The chapter number is on the $x$-axis, and the topics are listed on the $y$-axis. Color indicates the frequency of each topic within a chapter, calculated as the base-2 logarithm of the number of text chunks assigned to that topic.}
    \label{fig:topic_heat}
\end{figure}

Visual inspection of the topic clusters across chapters reveals notable shifts in thematic content, particularly around Chapters~11 and~26, as displayed in Figure~\ref{fig:topic_heat}. These shifts correspond to key narrative transitions within the book: Chapter~11 marks the beginning of the Hunger Games, the deadly battle to be the last person standing, and Chapter~26 signifies their conclusion.

\subsection{Description of the Extracted Topic Network}
\begin{figure}[htb!]
    \centering
    \includegraphics[width=.75\linewidth]{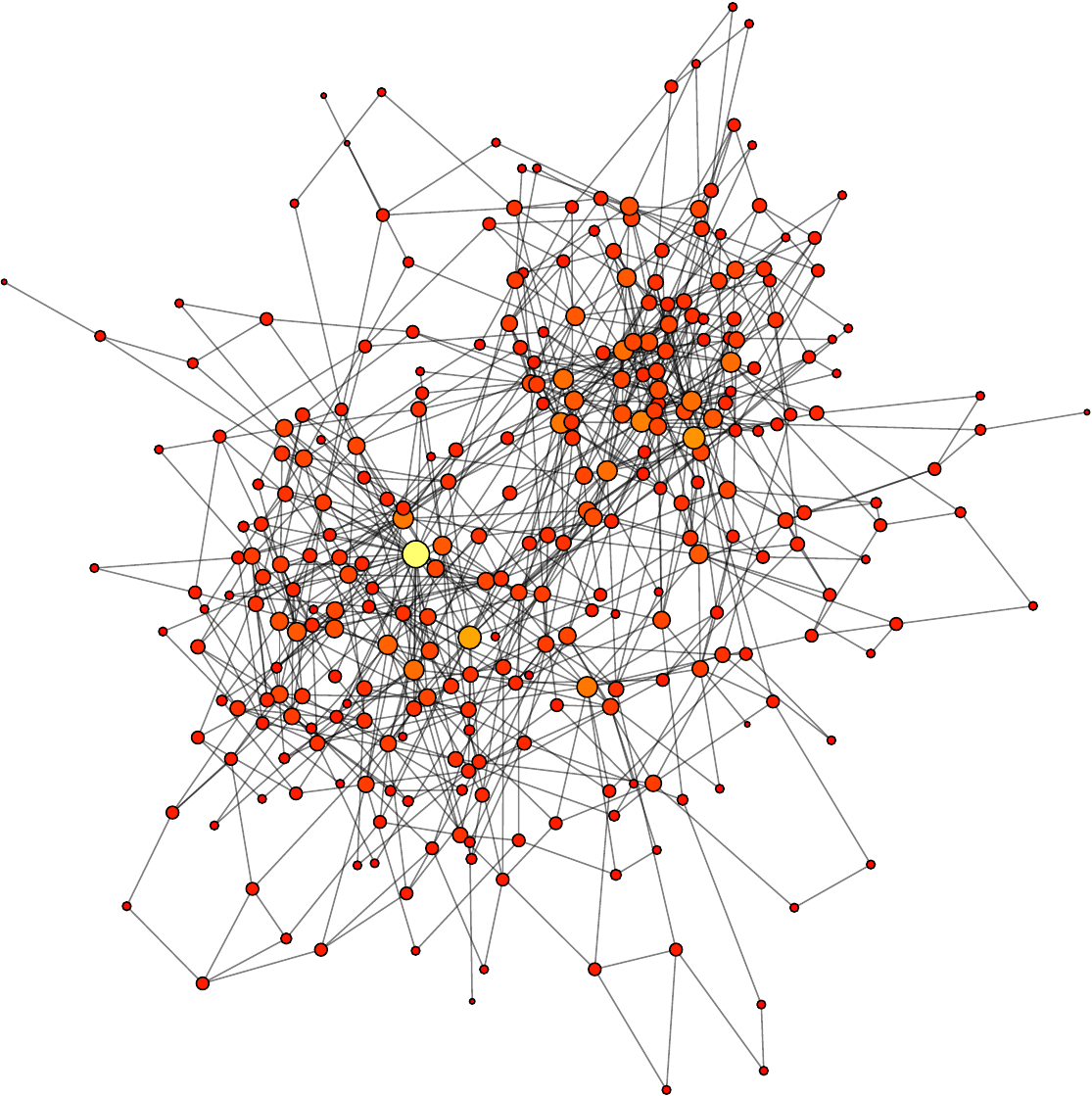}
    \caption{Topic network. The radius and the color of the vertices are proportional to their degree. The thickness of the edges is proportional to the cosine distance between the topic embeddings.}
    \label{fig:network}
\end{figure}

Figure~\ref{fig:network} shows the last graph constituting our dynamic network, which contains all topics and relationships for the whole novel. It consists of 302 vertices, representing the 302 identified topics, and 778 weighted undirected edges. It exhibits an average degree of 5.15 (SD = 3.13, median = 4, MAD = 3.00, range = 1--31), a weighted diameter of 1.49 (8 when considering unweighted edges), and its degree distribution follows a log-normal law (meanlog = 1.48, SDlog = 0.56). 

The small-worldness index, as defined by \citet{smallworld_humphries2008}, is 3.40, indicating a small-world network structure. The average shortest path length (unweighted) is 3.80. A significant hierarchical tendency is observed (hierarchical clustering coefficient = 0.16, $p < .001$; \citealp{HierarchyMeasureComplex_Mones2012}).

Community detection, using the Walktrap algorithm~\cite{walktrap_Pons2006}, which showed good performance in previous work~\cite{Yang2016}, reveals a minimum of two distinct communities within the network. A qualitative analysis of these communities reveals a strong correspondence with the narrative structure: one community primarily encompasses topics from Chapters~11--25 (the Hunger Games phase), while the other represents topics from the remaining chapters. 

\subsection{Persistent Homology}
\begin{figure}[htb!]
    \centering
    \includegraphics[width=1\linewidth]{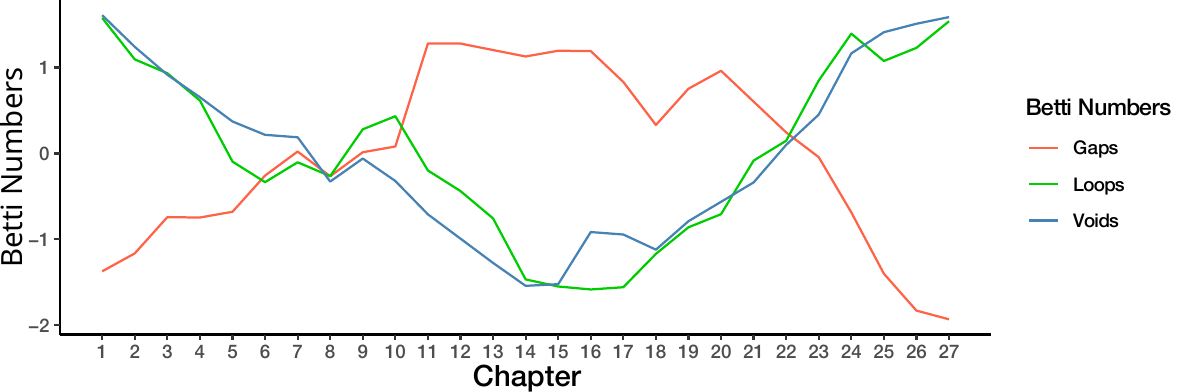}\\
    \includegraphics[width=1\linewidth]{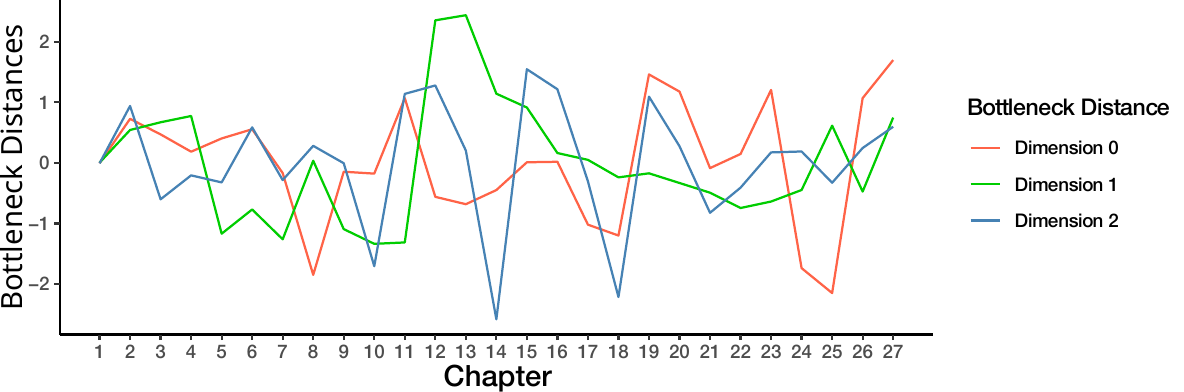}\\
    \includegraphics[width=1\linewidth]{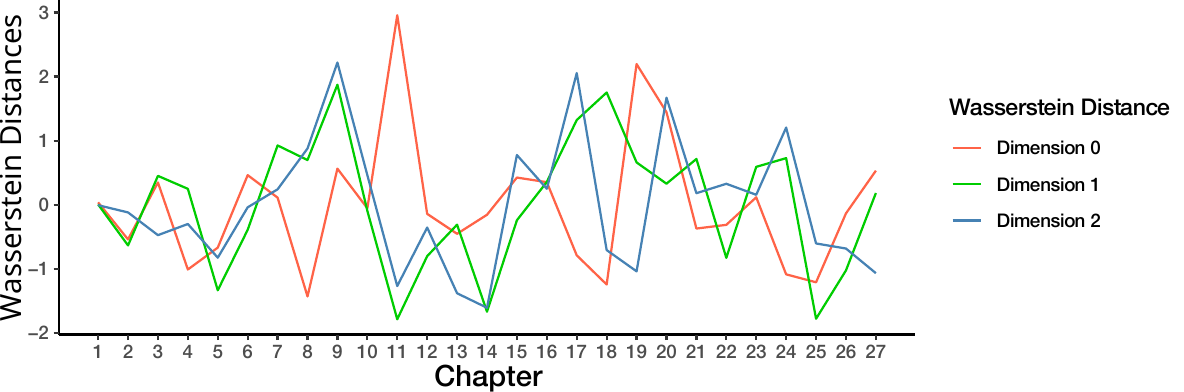}
    \caption{Total persistence feature counts ($\beta_0, \beta_1, \beta_2$; labeled ``Betti Numbers'' in the plot) per chapter (top), Bottleneck distances between chapters (middle), and Wasserstein distances between chapters (bottom). All measures are detrended.}
    \label{fig:PersistStats}
\end{figure}

All derived topological features are detrended, and are shown in Figure~\ref{fig:PersistStats}. The trends observed in the three plots exhibit a resemblance to the overarching narrative structure of ``The Hunger Games'', where the Games themselves commence in Chapter 11 and conclude in Chapter 25. Additional descriptive statistics are shown in Table~\ref{tab:descriptive_stats}.

\begin{table}
    \centering
    \begin{tabular}{lcccc}
        \hline
        \multicolumn{1}{l}{\textbf{Variable}} & \multicolumn{1}{c}{\textbf{Mean}} & \multicolumn{1}{c}{\textbf{SD}} & \multicolumn{1}{c}{\textbf{Min}} & \multicolumn{1}{c}{\textbf{Max}} \\
        \hline
        Curiosity & 69.5 & 3.9 & 60.3 & 77.0 \\
        Novel topics & 11.2 & 6.3 & 1.0 & 28.0 \\
        $\beta_0$ & 187.0 & 88.9 & 24.0 & 301.0 \\
        $\beta_1$ & 165.7 & 112.3 & 4.0 & 368.0 \\
        $\beta_2$ & 202.6 & 176.0 & 3.0 & 555.0 \\
        $\dist_B(\beta_0)$ & 0.4 & 0.0 & 0.3 & 0.5 \\
        $\dist_B(\beta_1)$ & 0.5 & 0.2 & 0.3 & 1.0 \\
        $\dist_B(\beta_2)$ & 0.3 & 0.1 & 0.2 & 1.0 \\
        $\dist_W(\beta_0)$ & 4.5 & 2.5 & 0.5 & 11.6 \\
        $\dist_W(\beta_1)$ & 6.2 & 2.1 & 2.4 & 9.8 \\
        $\dist_W(\beta_2)$ & 3.2 & 1.9 & 0.5 & 6.4 \\
        \hline
    \end{tabular}
    \caption{Descriptive statistics. $\beta_0$, $\beta_1$, and $\beta_2$ denote the total counts of finite persistence intervals extracted across the filtration for dimensions 0, 1, and 2. The Bottleneck and Wasserstein distances are denoted by $\dist_B$ and $\dist_W$.}
    \label{tab:descriptive_stats}
\end{table}

\subsection{Generalized Additive Model}
\begin{figure}[htb!]
    \centering
    \includegraphics[width=1\linewidth]{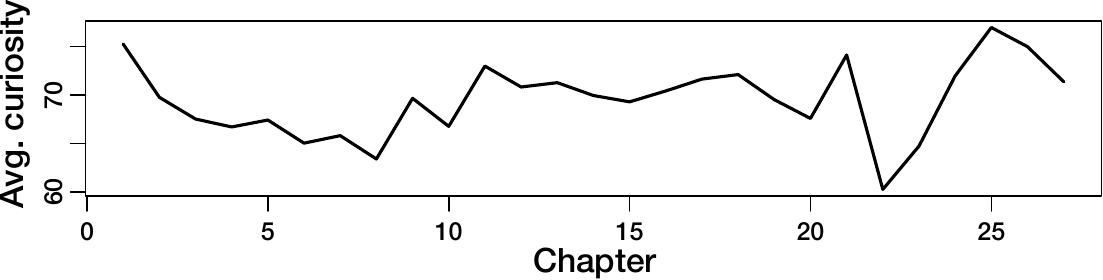}
    \caption{Average curiosity per chapter.}
    \label{fig:curiosity}
\end{figure}

\begin{figure}[htb!]
    \centering
    \includegraphics[width=1\linewidth]{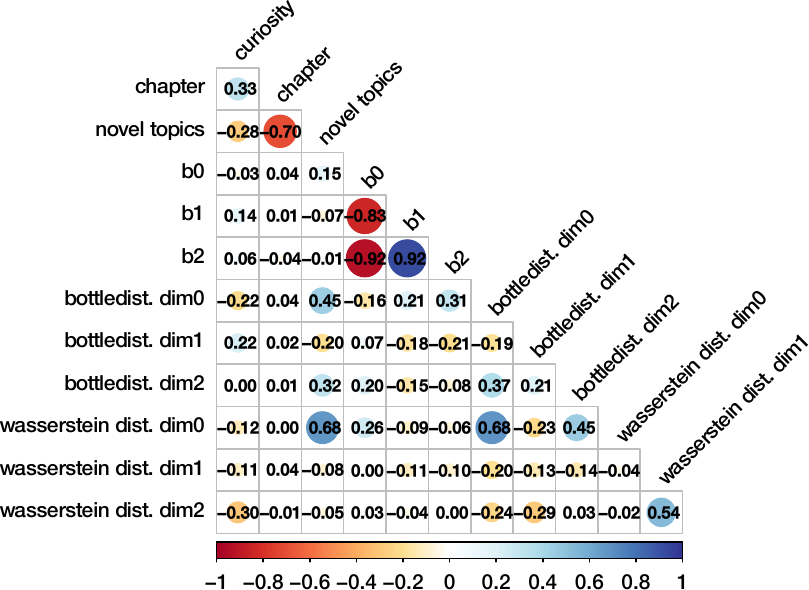}
    \caption{Spearman's bivariate rank correlations. Circle size and color are proportional to the correlation coefficient magnitude.}
    \label{fig:corre}
\end{figure}

The dependent variable in the generalized additive model (GAM) is the average curiosity score per chapter (M = 69.5, SD = 3.9, range: 60.3--77.0), as shown in Figure~\ref{fig:curiosity}. Figure~\ref{fig:corre} presents the bivariate Spearman's rank correlation matrix for all variables~\cite{spearman1904}.

We compare two GAMs to assess the unique contribution of topological features towards modeling the readers' curiosity:

\paragraph{Null Model (Control Variables only)}: It explains 23.8\% of the variance and 29.7\% of the deviance (permutation tests, both $p < .001$). 

\paragraph{Full Model (Topological Features + Control Variables)}: It explains 65.7\% of the variance (permutation test, $p < .05$) and 72.9\% of the deviance (permutation test, $p < .06$). 

A likelihood ratio test comparing the full model to the null model reveals a significant improvement in model fit ($\chi^2 = 11.25$, $df = 4.7$, $p < .001$). This indicates that the inclusion of topological features significantly improves the model's ability to explain variation in chapter-level curiosity, even after accounting for the number of novel topics and chapter index. These results show that our proposed pipeline, based on topological analysis of a dynamic topic network, can effectively model readers' curiosity ratings.


\section{Conclusion}
\label{sec:Conclu}
In this study, we combined existing methods into a pipeline for modeling semantic information gaps and explored their connection to reader curiosity, grounded in motivational psychology theories linking information gaps to curiosity. By constructing dynamic topic networks and using Topological Data Analysis to identify topological cavities as proxies for information gaps, we explored a distinct method from text structure analysis. Our preliminary findings demonstrate the feasibility of this approach, with statistical modeling indicating that topological features significantly enhance the prediction of reader curiosity beyond basic content and chapter progression. This proof-of-concept study provides a first step towards a quantifiable understanding of textual features associated with curiosity, potentially informing future research in computational text analysis and narrative understanding. Further research is needed to refine this pipeline, evaluate its generalizability across different text and media types, and investigate the specific contributions of various topological features to reader curiosity and engagement. A structured research agenda outlining methodological extensions, validation protocols, and narrative applications is provided in Appendix~\ref{sec:Agenda}.


\section*{Limitations}
First, our study includes a small sample size with limited measurement points focusing solely on the young adult novel ``The Hunger Games''. Because of this, and to avoid overfitting, we intentionally restricted our topological data analysis to simple summary measures, namely persistence feature counts and diagram distances. This may affect the robustness of our conclusions. Following this interdisciplinary proof-of-concept on one text, future work is essential to test the pipeline's robustness and applicability across diverse texts, genres (e.g., expository texts, news articles), and languages. 

Second, based on our relatively homogeneous UK-based sample, results may not directly generalize to different cultural contexts, where there can be significant cultural variations in reader response~\cite{CrossCulturalReader_Chesnokova2017}. The observed curiosity patterns are, thus, primarily reflective of this specific demographic group. Moderate reliability for mean curiosity ratings suggests that modeling individual reader differences is a key area for future work. 

Third, we model texts as topic networks rather than knowledge graphs, which could limit granularity. A possibility would be to explore semantic representation with LightRAG or GraphRAG~\cite{graphrag_Han2025, lightrag_guo2024}. 

Fourth, our network is undirected, which may overlook important learning dependencies, particularly in educational texts~\cite{knowlmap_Liu2012}. Future work could explore directed networks to better capture the sequential progression of knowledge. 

Fifth, we rely on shortest-path distances, but diffusion-based measures might better reflect how information evolves and interacts. 
Additionally, our focus on narrative structure constrains the applicability of our approach to raw embedding spaces. Investigating how to integrate narrative progression directly into embeddings, potentially through learning-theoretic models, should be explored. 
A sensitivity analysis exploring the impact of different embedding models and chunking parameters is provided in Appendix~\ref{sec:Sens}.
Though focusing on semantic/topological features was a purposeful choice for this study, the integration of linguistic features would be a valuable direction for future research. Finally, since our analysis is based on a narrative text, further research is needed to assess its effectiveness across different genres and domains. 

%

\bibliography{z3_Hoppetal2025_R3}

\appendix

\section{Sensitivity Analysis}
\label{sec:Sens}
To assess the robustness of our findings, we conducted a sensitivity analysis examining the influence of two key methodological choices: 1) the text embedding model and 2) the text segmentation parameters (window and overlap size). This analysis ensures that our primary conclusion---that topological features of a dynamic topic network significantly predict reader curiosity---is not an artifact of a specific parameterization but holds across different analytical conditions.

For this analysis, we re-ran our entire pipeline using two alternative state-of-the-art embedding models: Multilingual E5 Text Embeddings Large Instruct~\cite{Wang_E5embedding_2024} and Qwen3 Embedding 0.6B~\cite{Thang_qwen3embedding_2025}, both using 1,024 embedding dimensions. We compared their performance against our primary model, Voyage 3 Large~\cite{VoyageAI2025}, across several segmentation configurations: a larger window (w = 10, o = 3), medium windows (w = 7, o = 2, and w = 5, o = 2), and a smaller window (w = 3, o = 1). The rest of the analytical pipeline, including UMAP dimensionality reduction, HDBSCAN clustering, and GAM modeling, remained consistent with the main analysis. It is important to note that for the w = 3, o = 1 configuration, which produces very short text segments, the minimum cluster size for HDBSCAN was increased from 3 to 4 to ensure the formation of meaningful topics.

The comprehensive results, including network metrics and model fit statistics, are presented in Table \ref{tab:combined_net_metrics}. A focused summary comparing the deviance explained by the full and null models across all conditions is provided in Table \ref{tab:combined_model_perf}.

\begin{table}[h!]
    \centering
    \begin{tabular}{lccc}
        \hline
        \textbf{W/O} & \textbf{Topics} & \shortstack{\textbf{Wtd.} \\ \textbf{Diam.}} & \shortstack{\textbf{Unwtd.} \\ \textbf{Diam.}} \\
        \hline
        \multicolumn{4}{l}{\textit{Voyage 3 Large}} \\
        5/2 (Paper) & 302 & 1.49 & 8 \\
        10/3 & 121 & 1.20 & 7 \\
        7/2 & 179 & 1.60 & 8 \\
        3/1 & 280 & 1.07 & 5 \\
        \hline
        \multicolumn{4}{l}{\textit{E5 Large Instruct}} \\
        10/3 & 131 & 0.63 & 7 \\
        7/2 & 165 & 0.58 & 6 \\
        5/2 & 271 & 0.66 & 7 \\
        3/1 & 200 & 0.37 & 4 \\
        \hline
        \multicolumn{4}{l}{\textit{Qwen3 0.6B}} \\
        10/3 & 125 & 2.33 & 5 \\
        7/2 & 183 & 2.29 & 6 \\
        5/2 & 284 & 3.21 & 6 \\
        3/1 & 181 & 1.94 & 4 \\
        \hline
    \end{tabular}
    \caption{Network Metrics across Embedding Models. Abbreviations: W/O = Window/Overlap, Wtd. Diam. = Weighted Diameter, Unwtd. Diam. = Unweighted Diameter.}
    \label{tab:combined_net_metrics}
\end{table}

\begin{table}[h!]
    \centering
    \begin{tabular}{lccc}
        \hline
        \textbf{W/O} & \shortstack{\textbf{Full Dev.} \\ \textbf{(\%)}} & \shortstack{\textbf{Null Dev.} \\ \textbf{(\%)}} & \shortstack{\textbf{Full Dev.} \\ \textbf{p-val}} \\
        \hline
        \multicolumn{4}{l}{\textit{Voyage 3 Large}} \\
        5/2 (Paper) & 72.9 & 29.7 & $<$0.06 \\
        10/3 & 45.2 & 27.6 & 0.288 \\
        7/2 & 74.4 & 19.0 & 0.036 \\
        3/1 & 67.3 & 7.49 & 0.075 \\
        \hline
        \multicolumn{4}{l}{\textit{E5 Large Instruct}} \\
        10/3 & 64.6 & 7.49 & 0.069 \\
        7/2 & 28.6 & 7.49 & 0.418 \\
        5/2 & 84.3 & 7.49 & 0.017 \\
        3/1 & 75.8 & 7.49 & 0.022 \\
        \hline
        \multicolumn{4}{l}{\textit{Qwen3 0.6B}} \\
        10/3 & 78.8 & 22.7 & 0.032 \\
        7/2 & 68.5 & 7.49 & 0.073 \\
        5/2 & 88.1 & 7.49 & 0.016 \\
        3/1 & 74.6 & 7.49 & 0.039 \\
        \hline
    \end{tabular}
    \caption{Model Performance across Embedding Models. Abbreviations: W/O = Window/Overlap, Dev. = Deviance, p-val = Permutation Test p-value.}
    \label{tab:combined_model_perf}
\end{table}

\subsection{Results}
\label{sec:sensRes}
The sensitivity analysis confirms that the central finding of our study demonstrates considerable robustness. As shown in Table~\ref{tab:combined_model_perf}, the inclusion of topological features led to a statistically significant improvement in model fit compared to the null model across all tested embedding models and segmentation parameters (likelihood ratio tests, all $p < .05$), with permutation tests for full-model explained deviance reaching statistical significance ($p < .05$) in 6 of the 12 configurations and marginal significance ($p < .08$) in 4 additional configurations. This indicates that the relationship between the topological structure of the text and reader curiosity is a stable phenomenon.

Both the E5 and Qwen3 models proved to be viable alternatives, often producing models with high explanatory power. In particular, the Qwen3 model with a window of 5 and an overlap of 2 (w = 5, o = 2) achieved the highest explained deviance (88.1\%) of all tested configurations, closely followed by the E5 model with the same parameters (84.3\%). This suggests that the approach is effective with different modern embedding architectures.

The choice of window and overlap size influences the results, though no single configuration was universally optimal across all embeddings. Larger windows (e.g., w = 10) tend to identify fewer, broader topics, while smaller windows (e.g., w = 3) produce a more granular topic structure. The configurations using a window size of 5 or 7 sentences appear to strike an effective balance, frequently resulting in models with high predictive power. Although some alternative configurations resulted in slightly higher explained deviance, the performance of our primary model (Voyage 3 Large, w = 5, o = 2, 72.9\% explained deviance) is comparable to the best performing alternatives and is strongly supported by this analysis.

In summary, the sensitivity analysis confirms that our pipeline is not narrowly tuned to a single set of parameters. The ability of topological features to predict reader curiosity is a robust phenomenon, observable across different state-of-the-art embedding models and reasonable variations in text segmentation. This strengthens our confidence in the primary conclusions of the article and suggests the generalizability of the proposed method.

\section{A Short Introduction to Topological Data Analysis and Persistent Homology}
\label{sec:IntrTDA}

This section provides a brief, non-technical introduction to the concepts from Topological Data Analysis (TDA) used in our study, based on~\citet{TDA_Munch2017} and~\citet{Chazal_IntroductionTopologicalData_2021}. Our aim is to equip unfamiliar readers with the foundational understanding necessary to interpret how our topological measures of ``information gaps'' were derived from the textual data.

\subsection{Understanding Data Shape with Topology}
Topological Data Analysis (TDA) is a modern field that leverages principles from algebraic topology to discern the intrinsic ``shape'' or ``structure'' of data. Unlike traditional methods that might focus on local metrics or individual data points, TDA aims to identify robust global features, such as clusters, loops, and voids (see Figure \ref{fig:gaps}). The central tool within TDA for this purpose is \textit{persistent homology}.

To apply topological methods, data must first be represented in a suitable geometric form, typically as a \textit{simplicial complex}. A simplicial complex is a collection of fundamental building blocks called \textit{simplices}, which are analogous to points, line segments, filled triangles, and filled tetrahedra, and their higher-dimensional equivalents. Specifically, a \textit{0-simplex} is a single point (a vertex), a \textit{1-simplex} is an edge connecting two vertices, a \textit{2-simplex} is a filled triangle defined by three mutually connected vertices, and so on. In our work, the topics identified in the text serve as the fundamental points, or 0-simplices, of our topological analysis.

\subsection{Constructing Topological Spaces: The Vietoris--Rips Filtration}
Persistent homology does not analyze a static, singular complex; instead, it examines how the topological features of a complex evolve across a \textit{filtration}. A filtration is a nested sequence of simplicial complexes, where simplices are progressively added based on an increasing distance threshold, denoted by $\epsilon$. This process allows us to observe the data's structure at multiple scales, from fine-grained local connections to broader, global architectures.

For our dynamic topic networks, we use a specific type of filtration known as the \textit{Vietoris--Rips (VR) filtration}. This filtration is constructed upon a metric space where the points are the vertices of our topic network. The distance between any two vertices in this metric space is defined as the \textit{shortest path distance} between them within the graph. As the distance threshold $\epsilon$ increases, the VR filtration builds its complex: a $k$-simplex (a group of $k+1$ vertices) is added to the complex if the shortest path distance between \textit{every pair} of its constituent vertices is less than or equal to $\epsilon$. This means that if three topics are all mutually reachable within $\epsilon$ steps in the network, they will form a 2-simplex (a triangle), effectively ``filling in'' the space between them. 

\begin{figure}
    \centering
    \includegraphics[width=1\linewidth]{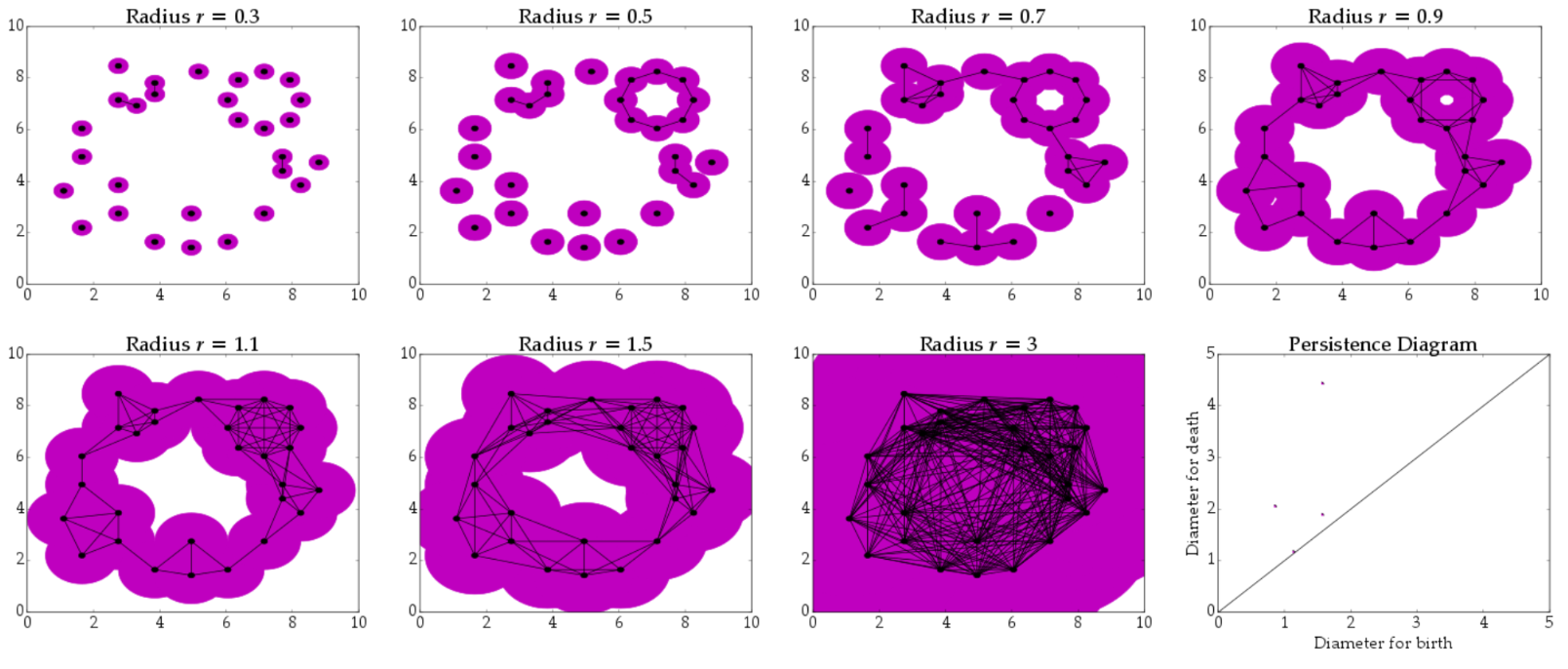}
    \caption{An illustration of a filtration and its corresponding persistence diagram. A point cloud (top left) is analyzed by progressively adding connections as a distance parameter \texttt{r} (radius) increases. This reveals a prominent circular feature (a loop) that persists across multiple scales. The persistence diagram (bottom right) summarizes this process, with the point far from the diagonal representing the robust, significant loop, and points close to the diagonal representing topological noise. Figure from~\citet{TDA_Munch2017}.}
    \label{fig:filtration}
\end{figure}

\textbf{Figure \ref{fig:filtration}} illustrates a filtration on a point cloud sampled from a circle. As the proximity parameter increases (shown as an expanding radius \texttt{r}, where the distance threshold $\epsilon = 2r$), more edges and triangles are added to the complex. Initially, only disconnected points exist. As the radius grows, local connections form, creating many small, transient loops. Crucially, a large, central loop emerges that reflects the circular shape of the data. This feature persists for a significant range of the filtration before the complex becomes dense enough to ``fill it in.''

\subsection{Quantifying Gaps: Betti Numbers and Persistence}

As filtration progresses and simplices are added, we use \textit{homology} to count the topological characteristics present at each value of $\epsilon$. These counts are formalized by \textit{Betti numbers} ($\beta_k$), which quantify the number of $k$-dimensional ``holes'' or ``gaps'' within the complex (see Figure \ref{fig:gaps}). In the context of our topic networks, these topological gaps offer interpretations regarding the information flow and thematic coherence:

\begin{itemize}
    \item \textbf{$\beta_0$ (Betti-0)}: Measures the number of \textit{connected components}. A high $\beta_0$ corresponds to many distinct topics. This can be interpreted as a thematic ``gap'' or disjunction in the narrative.
    \item \textbf{$\beta_1$ (Betti-1)}: Counts the number of one-dimensional \textit{loops or cycles}. These represent sequences of topics that form a closed cycle of relationships without being ``filled in'' by a higher-dimensional simplex. Such loops can indicate recurring themes, conceptual circularity in the narrative or ``plot holes'' that might provoke curiosity.
    \item \textbf{$\beta_2$ (Betti-2)}: Quantifies the number of two-dimensional \textit{voids or cavities}. These are hollow, sphere-like structures formed by collections of triangles that enclose an empty space. In our context, $\beta_2$ features could represent complex thematic ``plot holes'' where topics form a shell around an absent or implicit concept, potentially indicating a significant information gap.
\end{itemize}

The core idea of \textit{persistent homology} is to track the ``birth'' (emergence) and ``death'' (disappearance or filling) of these topological features across the entire filtration. For instance, a loop ($\beta_1$ feature) is born when its final edge completes the cycle and dies when a triangle is added that fills the loop. The \textit{persistence} of a feature is defined as the difference between its death and birth $\epsilon$ values ($\epsilon_{\text{death}} - \epsilon_{\text{birth}}$). While features with long persistence are typically considered indicative of stable structure, for this initial proof-of-concept study we focused on the total count of finite persistence intervals born across the entire filtration up to the network diameter (denoted $\beta_0, \beta_1, \beta_2$), rather than evaluating instantaneous Betti slices at a single $\epsilon$ or filtering features by lifespan. This choice was also informed by the preceding HDBSCAN clustering step, which already removes noisy embeddings and improves the quality of the unfiltered persistence feature counts. 
The results of persistent homology are typically visualized in a \textit{persistence diagram} (Figure \ref{fig:persist} and \ref{fig:filtration}), which plots each feature as a point indicating its birth and death times. By measuring distances between these persistence diagrams (e.g., using Wasserstein or Bottleneck distances), we can quantify the degree of structural change in the topic network between consecutive text segments, thereby providing a computational proxy for the evolving information gaps that drive reader curiosity.

For a more detailed introduction to TDA, we recommend the work by~\citet{TDA_Munch2017}, \citet{TDA_Intro_Carlsson2020}, and~\citet{Chazal_IntroductionTopologicalData_2021}.

\section{Inter-Rater Reliability}
\label{sec:Icc}
In order to assess Inter-Rater Reliability for mean chapter curiosity (cf. Section~\ref{sec:SetupPrepr}), we used the following formula by \citet{Shrout1979}:
\begin{equation}
    ICC = \frac{\sigma^{2}_{c}}{\sigma^{2}_{c}+ \sigma^{2}_{r, e}/k },    
\end{equation}
where $\sigma^2_{c}$ is the variance between chapters, $\sigma^2_{r, e}$ residual variance (which includes the rater-by-chapter interaction and measurement error), and $k$ is the number of raters.

A linear mixed-effects model was fitted to predict curiosity scores, including random intercepts for both chapter and subject. This model served to estimate the necessary variance components for the reliability assessment. The analysis, based on 1323 observations collected from 27 distinct chapters and 49 subjects, resulted in an estimated overall mean curiosity score (Intercept) of 69.54 (SE = 2.29, 95\% CI [65.05, 74.03]), which was statistically significant ($p < .001$). The estimated variance components were: variance between chapters ($\sigma^2_{\text{chapter}}$) = 10.48, variance between subjects ($\sigma^2_{\text{subject}}$) = 229.95, and residual variance ($\sigma^2_{\text{residual}}$) = 214.36. 

Specifically, to assess the reliability of the aggregated curiosity scores for distinguishing between chapters, with $k=49$ participants serving as raters, an ICC of 0.71 was calculated using the estimated chapter variance component ($\sigma^2_{c} = 10.48$) and the residual variance ($\sigma^2_{r, e} = 214.36$). The calculated reliability of 0.71 for aggregated curiosity scores across 49 participants, while acceptable, indicates a moderate to limited ability to robustly distinguish between chapters. Achieving higher precision in differentiating chapters might benefit from a larger sample of participants.

\section{Research Agenda for Narrative Topology}
\label{sec:Agenda}
This appendix outlines concrete next steps to extend and validate the narrative topology framework presented in the main text. We organize this agenda into four areas: graph extraction and validation, topic and narrative analysis, temporal story structures, and visualization and software tooling.

\subsection{Graph Extraction and Model Validation}
To test whether topological signals reflect genuine narrative progression rather than statistical artifacts of topic co-occurrence, future work should introduce configuration null models that randomize text chunks before constructing the topic network. Comparing original and randomized networks will show how sequence disruption alters persistent homology summaries and permutation-test outcomes. Partial randomization, where only selected passages or chapter orders are shuffled, can further test how incremental losses of narrative coherence affect model performance. In addition, the current graph connects only consecutive chunks; extending edge construction to link non-consecutive but semantically close topics in embedding space, using distance scaling or linear combinations of sequential and semantic proximity, could capture long-range thematic callbacks.

\subsection{Topic Analysis and Narrative Archetypes}
Understanding what each topological gap represents requires linking structural metrics to literary content. First, extracted topics can be grouped into functional archetypes, such as character-oriented, lore, environment, and action topics, to examine which thematic roles generate persistent cycles and voids. Second, inspecting typical cycle lengths in dimension~1 ($H_1$) will help establish empirical bounds on narrative gaps in story flow. Third, comparing mystery-centric genres (such as crime fiction) against non-mystery texts (such as biographies), alongside manual validation on stories chosen for distinctive structural traits, will test how known narrative properties shape graph metrics and homology features. Fourth, moving beyond chapter means to examine the full distribution of reader curiosity scores across chapters will clarify how topological patterns relate to reader-level variation.

\subsection{Story Types and Temporal Structure}
Because \textit{The Hunger Games} follows a largely linear timeline, extending the pipeline to non-linear narratives is an important test of generality. Future studies should examine how flashbacks and other non-chronological devices alter dynamic topic graphs and persistent homology measures. In complex multi-timeline works such as \textit{A Song of Ice and Fire}, comparing graphs ordered by diegetic (in-story) time against graphs ordered by reading order can quantify narrative time dilation. Similarly, analyzing repetitive or looping structures, such as \textit{Groundhog Day}, will test how recurring cycles and incremental plot variations appear in persistent homology diagrams.

\subsection{Visualizations, Monitoring, and Tooling}
Interpreting dynamic topological structures in long texts will benefit from dedicated visual diagnostics. Promising approaches include animated plots of consecutive topics in embedding space that reveal vertex and edge additions chunk by chunk or trace paths across the full graph, topical sequence maps that display an entire novel's topic progression as colored sub-rectangles encoding topical distance per chunk, and chapter-level plots tracking average and maximum persistence lifetimes alongside feature counts. Combined with reader feedback, these diagnostics can help quantify the effectiveness of specific writing devices such as voluntary information hiding. Finally, consolidating this workflow into an open-source R package would make narrative topology methods accessible for broader computational literary studies.


\end{document}